\documentclass[conference]{IEEEtran}
\usepackage{booktabs}
\usepackage{multirow}
\IEEEoverridecommandlockouts
\usepackage{cite}
\usepackage{hyperref}
\usepackage{amsmath,amssymb,amsfonts}
\usepackage{algorithmic}
\usepackage{graphicx}
\usepackage{textcomp}
\usepackage{xcolor}
\usepackage{pifont}
\usepackage{xspace}
\usepackage{bbm}
\newcommand{\proj}{\textsc{ZeroSim}\xspace}
\def\BibTeX{{\rm B\kern-.05em{\sc i\kern-.025em b}\kern-.08em
    T\kern-.1667em\lower.7ex\hbox{E}\kern-.125emX}}
\begin{document}

\title{\proj: Zero-Shot Analog Circuit Evaluation with Unified Transformer Embeddings}

\author{Xiaomeng Yang, Jian Gao, Yanzhi Wang, Xuan Zhang
\\ \textit{Department of Electrical and Computer Engineering, Northeastern University, Boston, USA} \\ Email: \{yang.xiaome, gao.jian3, yanz.wang, xuan.zhang\}@northeastern.edu \\
}


\maketitle

\begin{abstract}
Although recent advancements in learning-based analog circuit design automation have tackled tasks such as topology generation, device sizing, and layout synthesis, efficient performance evaluation remains a major bottleneck. 
Traditional SPICE simulations are time-consuming, while existing machine learning methods often require topology-specific retraining or manual substructure segmentation for fine-tuning, hindering scalability and adaptability. In this work, we propose \textbf{\proj}, a transformer-based performance modeling framework designed to achieve robust in-distribution generalization across trained topologies under novel parameter configurations and zero-shot generalization to unseen topologies without any fine-tuning. 
We apply three key enabling strategies: (1) a diverse training corpus of 3.6 million instances covering over 60 amplifier topologies, (2) unified topology embeddings leveraging global-aware tokens and hierarchical attention to robustly generalize to novel circuits, and (3) a topology-conditioned parameter mapping approach that maintains consistent structural representations independent of parameter variations. Our experimental results demonstrate that \proj significantly outperforms baseline models such as multilayer perceptrons, graph neural networks and transformers, delivering accurate zero-shot predictions across different amplifier topologies. Additionally, when integrated into a reinforcement learning-based parameter optimization pipeline, \proj achieves a remarkable speedup ($13\times$) compared to conventional SPICE simulations, underscoring its practical value for a wide range of analog circuit design automation tasks.
\end{abstract}


\section{Introduction}
In recent years, large neural networks have revolutionized artificial intelligence. Foundation models such as GPT~\cite{brown2020language,achiam2023gpt}, CLIP~\cite{radford2021learning}, and SAM~\cite{kirillov2023segment}, have demonstrated exceptional capabilities across tasks ranging from language understanding to image segmentation, sparking a paradigm shift toward unified architectures adaptable to diverse applications with minimal additional retraining~\cite{bommasani2021opportunities}. This paradigm shift underscores the potential of unified and scalable approaches to streamline complex workflows and reduce manual intervention.

A similar AI transformation is underway in Electronic Design Automation (EDA).
As semiconductor technology advances and the complexity of integrated circuits continues to grow, conventional integrated circuit (IC) design methodologies have become increasingly labor-intensive, error-prone, and inefficient~\cite{ren2022machine}. In particular, analog circuit design remains a major bottleneck in the chip design process due to its manual and iterative nature. Designers must explore complex topologies, adjust device parameters through extensive simulation cycles, and balance conflicting performance metrics~\cite{mina2022review,ren2022machine}. Therefore, researchers have begun investigating automation methods to address these challenges effectively.

Recent advancements in AI-driven EDA focus on automating the entire analog circuit design workflow, encompassing topology generation, device sizing optimization, and layout automation. Graph-based and generative model-based techniques (e.g., CktGNN~\cite{dong2023cktgnn}, AnalogCoder~\cite{lai2024analogcoder}) have significantly advanced topology generation by encoding relationships between devices and inferring structural compositions. In device sizing optimization, reinforcement learning (RL) and Bayesian optimization methods (e.g., AutoCkt~\cite{settaluri2020autockt}, \cite{liu2020transfer}) provide efficient sequential decision-making guided by performance feedback. Layout automation has likewise benefited from frameworks that convert schematics into manufacturable layouts through hierarchical, constraint-aware models (e.g., ALIGN~\cite{kunal2019align,dhar2020align}).

Despite significant progress in these automation stages, rapid and accurate circuit performance evaluation remains a substantial bottleneck. Traditional SPICE-based simulations are computationally intensive and time-consuming, making them unsuitable for the iterative loops of an AI-driven design flow~\cite{gao2023rose}. There is an urgent need to integrate efficient, scalable circuit performance modeling methods that can evaluate circuits in real time and support analog circuit design automation~\cite{hassanpourghadi2021circuit}. Current learning-based evaluation methods often resort to training specialized, topology-specific models~\cite{poddar2024insight} or rely on transfer learning approaches that demand manual segmentation and fine-tuning on each sub-topology~\cite{wu2022transfer,fayazi2023funtom}. Although these methods can deliver modest accuracy, they are neither scalable nor adaptable. Introducing a new topology requires reconfiguration or retraining the models, incurring significant overhead.
These approaches typically call for large, labeled datasets for every new topology, making them costly to maintain and deploy in industry. An ideal evaluation model should support both in-distribution generalization, accurately predicting performance metrics for circuit topologies seen during training but with novel parameter configurations, and zero-shot generalization, reliably evaluating entirely unseen circuit topologies across diverse parameter settings without additional retraining or fine-tuning.

To overcome these limitations, we propose \textbf{\proj}, a transformer-based zero-shot performance modeling framework for analog circuit evaluation. Our objective is to create a unified model capable of addressing three critical challenges inherent in analog circuit design: (1) the combinatorial explosion of possible circuit topology and parameter settings, necessitating comprehensive representation generalization; (2) the complex and high-dimensional parameter space within each topology, requiring precise modeling to capture intricate performance relationships; and (3) ensuring structural representations remain consistent and robust, independent of parameter variations.

Our framework directly tackles these challenges through three corresponding strategies: (1) Diverse Training Corpus: We construct a sufficiently large-scale dataset containing 3.6 million circuit instances, each defined by a unique combination of amplifier topology and parameter settings, covering over 60 representative topologies including one-stage, two-stage, and three-stage designs with varying device numbers from 6 to 39. (2) Unified Topology Embeddings: We represent circuit structures through a single, unified embedding strategy utilizing global-aware tokens and hierarchical attention mechanisms. This unified representation explicitly captures fine-grained device-level connectivity patterns and contextual relationships simultaneously, enabling robust generalization to novel topologies. (3) Topology-Conditioned Parameter Mapping: We condition performance predictions on topology-specific embeddings learned independently of parameter information. This separation ensures that the learned structural representation remains consistent and robust, unaffected by variations in injected parameters, thus enabling effective generalization across unseen designs. Unlike methods that require topology-specific retraining or task-specific architectures, \proj is trained once and supports zero-shot performance prediction. This means it can be directly applied to new amplifier topologies and parameter settings without fine-tuning, enabling scalable and flexible design workflows.

Experimental results demonstrate that our model consistently outperforms traditional multilayer perceptrons and graph neural networks in both in-distribution and zero-shot evaluation settings. Moreover, by integrating \proj into a reinforcement learning-based sizing optimization pipeline, we achieve a 13\texttimes{} speedup in overall design time compared with traditional SPICE simulations, demonstrating its practical value in accelerating iterative analog design. The primary contributions of our work can be summarized as follows: 

\begin{itemize}
\item We collect a large, diverse dataset spanning 60 topologies and 3.6 million instances with varying design complexity.
\item We introduce \proj, a global-knowledge aware transformer for zero-shot performance modeling, generalizing to unseen amplifier topologies without retraining.
\item We propose a unified embedding strategy that encodes both topology and device parameters, capturing essential structural patterns and enabling scalable generalization beyond fixed designs.
\item Through extensive experiments, we demonstrate that \proj significantly outperforms existing learning-based baselines and accelerate the RL optimization compared with SPICE simulation.
\end{itemize}

\section{Background and Related Work}
\subsection{AI Driven Circuit Design}
Analog circuit design traditionally comprises three tightly coupled stages: designing a circuit topology that achieves the desired functionality, sizing individual devices to meet performance specifications, and producing a manufacturable layout. This process often involves extensive iterations and significant manual effort, which impacts overall productivity and increases the complexity of design tasks. As integrated circuits become increasingly complex, these challenges have motivated significant progress in AI-driven automation across the design pipeline~\cite{chen2024dawn}.

To alleviate the manual burden and inefficiency in analog design, AI-driven methods have been developed to automate these key stages. Circuit topology generation determines device interconnections and hierarchical structures. Graph-based approaches like CktGNN~\cite{dong2023cktgnn} and AnGeL~\cite{fayazi2023angel} leverage hierarchical graph neural networks and semi-supervised learning to encode reusable subgraphs. Meanwhile, generative models such as AnalogCoder~\cite{lai2024analogcoder}, LaMAGIC~\cite{chang2024lamagic}, Artisan~\cite{chen2024artisan}, and AnalogGenie~\cite{gao2025analoggenie} employ large language models, autoregressive decoding, and multi-agent strategies to synthesize circuit topologies or design code. 
For device sizing, reinforcement learning (RL) and Bayesian optimization have proven effective in navigating complex design spaces. For instance, AutoCkt~\cite{settaluri2020autockt} treats transistor sizing as a sequential decision-making process guided by performance feedback, while the GCN-RL circuit designer~\cite{wang2020gcn} uses graph convolutional networks (GCN) to enhance generalization across multiple topologies. Other frameworks, such as RoSE~\cite{gao2023rose} and RoSE-Opt~\cite{cao2024rose}, incorporate domain-specific knowledge to improve both accuracy and efficiency.
Layout automation has also benefited from machine learning. ALIGN~\cite{kunal2019align,dhar2020align} uses constraint-aware layout synthesis via hierarchical analysis, and MAGICAL~\cite{xu2019magical} further integrates expert inspired design rules for automated placement and routing with minimal manual intervention.

Despite these advances, performance evaluation remains a crucial bottleneck of AI-driven analog design. Each candidate design of a new topology or sizing configuration must be evaluated through simulation to provide feedback for subsequent iterations. This evaluation step is revisited hundreds to thousands of times in a typical design loop, making it a dominant contributor to overall runtime. Traditionally, the evaluation process relies on SPICE simulations~\cite{gao2023rose}, which are computationally expensive and time-consuming. Consequently, there is an increasing need for developing efficient performance evaluation models that can predict comprehensive circuit behavior far faster than classic SPICE simulations, especially when dealing with new topologies or parameter configurations.

\subsection{Learning-based Circuit Evaluation}
Early learning-based performance predictors relied on simple feedforward networks (MLPs) that treat circuits as flat vectors of device parameters~\cite{liu2021specification}. While easy to implement, these models failed to capture structural relationships among components, limiting their ability to generalize across topologies. Hybrid approaches such as FuNToM~\cite{fayazi2023funtom} attempted to improve scalability by using two-port MLPs for compositional modeling of RF circuits. However, the reliance on topology-specific structures constrained their applicability to broader design spaces. Graph neural networks (GNNs) brought significant advances by explicitly encoding circuit structure.
ParaGraph~\cite{ren2020paragraph} represents circuits as heterogeneous graphs to predict parasitics and device properties, outperforming traditional estimators. GCN-RL Circuit Designer~\cite{wang2020gcn} integrates GNN-based Circuit Evaluation with RL. To enhance generalization, Hakhamaneshi et al.~\cite{hakhamaneshi2022pretraining} pretrained GNNs on large circuit datasets and demonstrated their few-shot adaptation capabilities. Wu and Savidis~\cite{wu2022transfer} further explored transferring GNN-based performance models across multiple circuits.

Moving beyond pure GNNs, transformer-based architectures have recently gained attention for their expressiveness and scalability. Fan et al.~\cite{fan2024graph} introduce a graph-transformer network tailored to converter circuits, achieving efficient evaluation across different topologies. AICircuit\cite{mehradfar2024aicircuit} systematically benchmarks backbone models and confirms the superiority of transformer-based surrogates in modeling nonlinear performance metrics. Expanding further, INSIGHT~\cite{poddar2024insight} treats analog simulation as an autoregressive task, using sequential modeling to predict detailed performance curves in microseconds but it can only be applied for a certain circuit topology.

These methods highlight the evolution from parameter-only surrogates to topology-aware models. However, existing solutions either (1) require per-topology retraining or fine-tuning, or (2) assume a relatively limited topology design space. In contrast, our proposed \proj offers a unified framework that jointly encodes structural and parametric information, enabling \emph{zero-shot performance prediction} across different circuit topologies without the need for specialized retraining or network customization.

\section{Method}
\begin{figure*}[!t]
\centering
\includegraphics[width=6in]{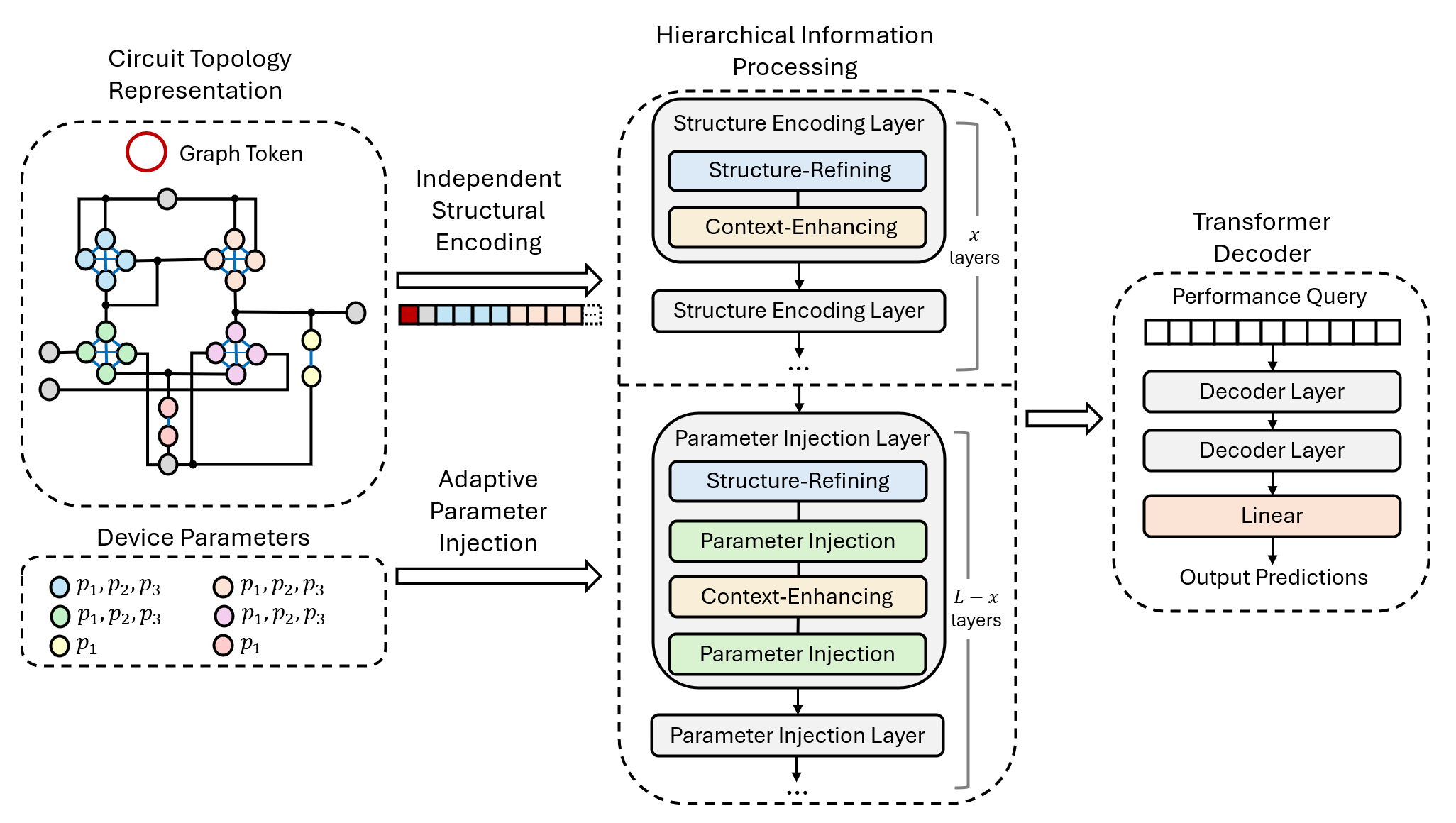}
\caption{The overall pipeline of \proj. It first encodes circuit topology using independent structural encoding with hierarchical attention. Device parameters are then injected via cross-attention in the parameter-aware layers. A transformer decoder processes performance query tokens to generate final circuit performance predictions.}
\label{fig:model}
\end{figure*}

In this section, we introduce \proj, a transformer-based zero-shot performance modeling framework for analog circuit simulation. 
Our goal is to develop a unified model that addresses the dual challenges inherent in analog design: the combinatorial explosion of possible circuit topologies and the continuous, high-dimensional parameter space associated with individual devices.
Even with a small library of basic building blocks, the number of valid topologies grows rapidly. For example, over 191k unique topologies can be generated using just 5 component instances~\cite{fan2024graph}, creating substantial challenges for both representation and generalization. Moreover, within any single topology, the device parameter space (e.g., transistor length L, width W, and multiplicity M; or resistance and capacitance values for passive components) remains vast and highly complex.

As shown in Fig.~\ref{fig:model}, \proj tackles these issues through two key design principles: (1) a hierarchical encoding process that simultaneously captures local connectivity patterns and global context information within the topology, and (2) a progressive knowledge fusion mechanism that introduces device parameters into the model without undermining the uniformity or expressiveness of the structural representation. We will describe each component of this framework in detail in the following subsections.

\subsection{Circuit Topology Representation}
\begin{figure}[!t]
\centering
\includegraphics[width=1\columnwidth]{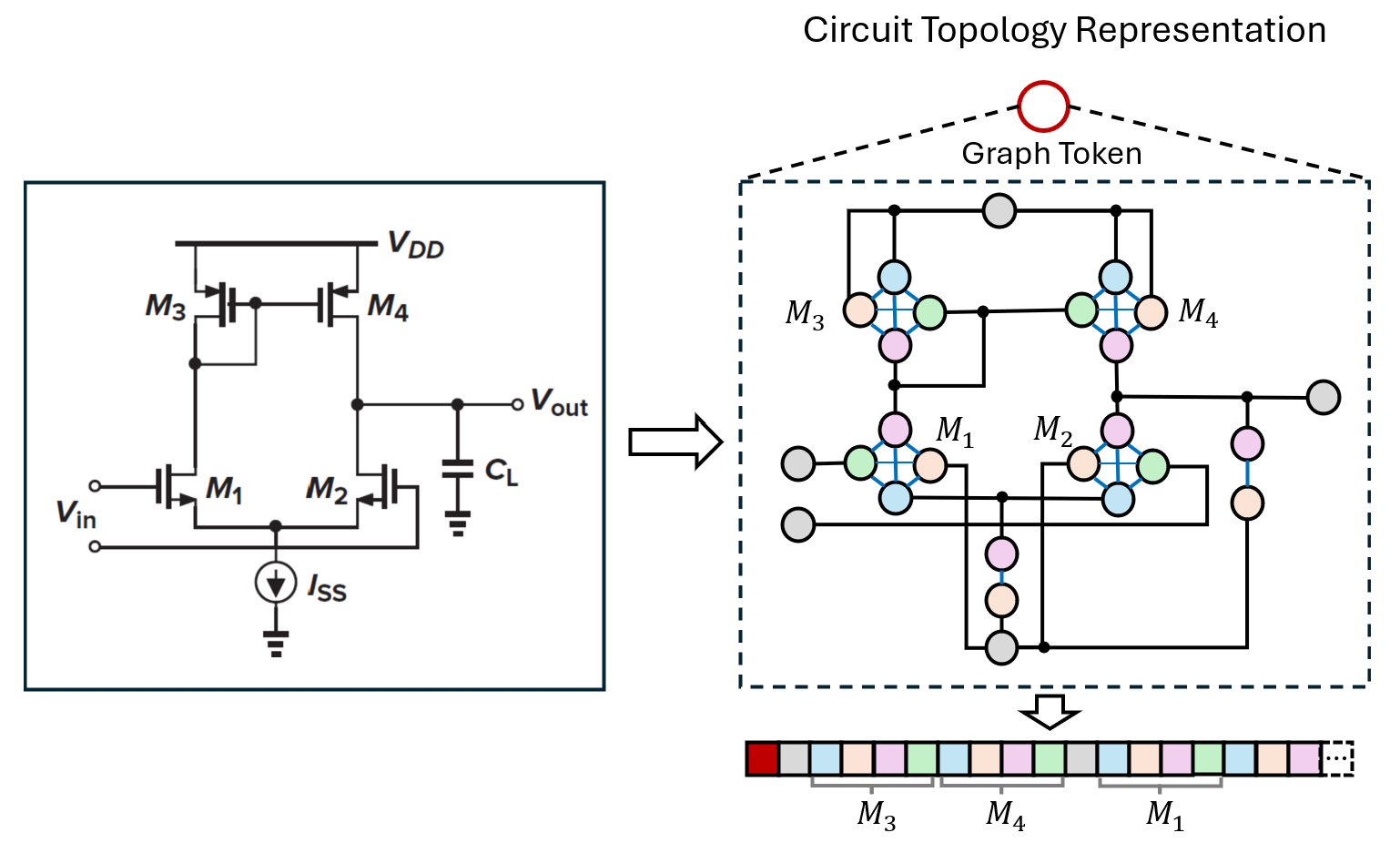}
\caption{The transformation of a transistor-level amplifier schematic into a graph-based topology representation. Each pin-level node and device is encoded into a structured graph, augmented with a global graph token to facilitate context-aware encoding.}
\label{fig:topology}
\end{figure}

To capture the structural nuances of analog circuits, we represent each schematic as a pin-level graph. Formally, let $G=(V,E)$ denotes the graph, where each node $v\in V$ corresponds to a pin of a circuit component and each edge $(v_i, v_j)\in E$ signifies that the respective pins are physically connected in the schematic. This pin-level representation is crucial because analog circuit behaviors often depend on specific pin connections (e.g., D, G, S, B for MOSFET) rather than on the device as a whole. Additionally, we insert virtual edges among all pins belonging to the same device to encode device-centric structural subgraphs. Even if a device has multiple pins, this strategy ensures that all relevant pins remain interconnected, allowing the model to reason about each device's local structure.

Beyond local connectivity, it is essential to introduce a global “summary” capable of capturing high-level circuit behaviors such as loops, feedback paths, and hierarchical groupings. As illustrated in Fig.~\ref{fig:topology}, we introduce a special token $[G]$ that represents the entire circuit:
\begin{equation}
[G] \longleftrightarrow \{v_1,v_2,...,v_{|V|}\}
\end{equation}
During the attention process, $G$ interacts with all node embeddings, effectively serving as a bridge between distant pins, devices, or subgraphs that may not be directly connected within local neighborhoods.

\subsection{Hierarchical Information Processing}
\label{sec:hierarchical}
A major challenge in encoding analog circuits lies in balancing detailed, device-level connectivity with broader circuit-wide interactions. \proj addresses this through a hierarchical, alternating attention mechanism that processes topology information at two complementary levels of granularity.

\subsubsection{Attention mechanism}
To begin, we briefly review the multi-head attention (MHA) operation. Given queries $Q$, keys $K$, and values $V$ of dimension $d$, the scaled dot-product attention is computed as:
\begin{equation}
Attn(Q, K, V) = softmax(\frac{QK^T+M}{\sqrt{d}})V,
\end{equation}
where $M$ is an optional additive mask used to restrict attention to certain positions. In the multi-head setting with $h$ heads, each head $i$ applies learned linear projection $W^Q_i, W^K_i, W^V_i$. The results from all heads are then concatenated and projected again: $MHA(Q,K,V)=Concat(h_1, ..., h_h)W^O, h_i = Attn(QW^Q_i, KW^K_i, VW^V_i)$.
Different masks $M$ can be applied within this framework to impose local or global attention, which is critical for capturing hierarchical information.

\subsubsection{Structure-level refining}
The first encoding phase focuses on capturing localized interactions within subgraphs that emphasize device-level connectivity and immediate electrical interactions. Let $H^{(l-1)}\in\mathbb{R}^{(N+1)\times d}$ be the hidden states from layer $l-1$, representing $N$ pin-level nodes plus the global token $[G]$. The structure-level encoder calculates:
\begin{equation}
\widetilde{H}^{(l)} = MHA(H^{(l-1)}, H^{(l-1)}, H^{(l-1)}; M_{local}),
\end{equation}
where $M_{local}$ restricts attention to each node itself, its immediate neighbors, and any pins within the same device subgraph. This mask enforces localized connectivity, enabling the model to learn fundamental physical constraints and small-scale interaction patterns essential for device-level accuracy.

\subsubsection{Context-level enhancing}
Subsequently, the next phase aims to capture broader context by allowing long-range attention across the graph. In this phase, each device can exchange information with more distant subcircuits:
\begin{equation}
\widetilde{H}^{(l)} = MHA(H^{(l-1)}, H^{(l-1)}, H^{(l-1)})
\end{equation}
which permits each node to attend to every other node, thereby accommodating complex multi-device relationships such as feedback loops.

By alternating between structure-level refining and context-level enhancing blocks for a total of $L$ layers, \proj is able to capture both detailed local interactions and circuit-wide behaviors. This encoding strategy is crucial for modeling the nuanced phenomena present in analog designs, ranging from local current mirrors to more complex global feedback loops.

\subsection{Parameter-Aware Progressive Knowledge Fusion}
While circuit topology defines the overall connectivity of an analog circuit, performance metrics also depend heavily on continuous-valued device parameters, such as transistor dimensions or resistor and capacitor values. A central challenge is to incorporate these parameters into the learned representation without undermining the zero-shot adaptability of the structural encoding. To address this, \proj adopts a two-stage pipeline: it first encodes topological relationships independently, then introduces device parameters through a device-specific cross-attention mechanism, guided by careful masking.

\subsubsection{Independent structural encoding}
In the initial phase, \proj focuses exclusively on learning the topological structure. Let $x$ be a hyperparameter indicating how many of the early transformer layers are devoted to structure-only encoding. Denote the hidden representations at layer $x$ by $H^{(x)}$. In the first $x$ layers, no parameter embeddings are introduced. Instead, the model relies solely on the pin-level graph and the global token $[G]$ to form a high-fidelity structural representation:
\begin{equation}
    H^{(x)} = f_{structure}(H^{(0)}),
\end{equation}
where $H^{(0)}=[[G],h_1^{(0)},...,h_N^{(0)}]$ includes initial embeddings of nodes each pin-level node plus the global token and $f_{structure}$ denotes the stack of structure and context-level transformer blocks (\ref{sec:hierarchical}). By postponing the introduction of parameter embeddings, this phase yields a reusable structural encoding that is agnostic to specific device settings.

\subsubsection{Adaptive parameter injection}
Once the structural backbone is established, \proj incorporates device parameters. Suppose the circuit contains $m$ distinct devices, each with a set of parameter values $\{p_{i1}, p_{i2}, ...\}$. We first embed each parameter $p_{ij}$ into a $d$-dimensional token: $r_{ij}=Embed_\theta(p_{ij})\in\mathbb{R}^d$, where $Embed_\theta$ is a simple linear layer or a small multi-layer perceptron (MLP). Then, each device $i$ is associated with a collection of parameter tokens $\{r_{i1}, r_{i2}, ...\}$.

In the subsequent transformer layers, each pin-level node $h^{(l-1)}_v$ performs cross-attention over the parameter tokens belonging to its corresponding device. Let $\mathcal{V}_i$ denote the set of nodes that belong to device $i$. For node $v\in\mathcal{V}_i$ is:
\begin{equation}
h_v^{(l)} = MHA(h^{(l-1)}_v, \mathcal{V}_i, \mathcal{V}_i; M_{device}),
\end{equation}
where the $M_{device}$ ensures that node 
v only attends to parameter tokens relevant to its device. This design prevents spurious parameter influences between unrelated parts of the circuit. The global token $[G]$ is allowed to attend to parameter tokens from devices, aggregating circuit-wide parametric information into a unified representation.

By separating the topological encoding from parameter injection, \proj preserves a consistent structural representation that can seamlessly adapt to new device configurations and even new topologies. This two-stage approach ensures the representation robustness, scalability, and zero-shot capability, aligning naturally with real-world analog design workflows.

\subsection{Performance Prediction}
We employ a shallow transformer decoder to produce the final set of performance predictions $\hat{y}\in\mathbb{R}^k$ based on $H^{(L)}$, where $k$ is the number of target performance metrics. We introduce a set of query tokens $Q_p = \{q_{p1}, ..., q_{pk}\}$, each corresponding to a particular performance metric. These query tokens attend to the encoder outputs: 
\begin{equation}
    Y=Decoder(Q_p, H^{(L)}, H^{(L)}),
\end{equation}
where $Y\in\mathbb{R}^{k\times d}$ is the hidden representation for each metric query. A final linear layer then maps $Y$ to the predicted performance values through $\hat{y}_i = W_{out}Y_i + b_{out}$ for $i=1,...,k$. 
This query-driven design naturally aligns with the multi-task nature of analog circuit performance prediction. It encourages the model to develop specialized sub-representations for each metric benefiting from shared topology–parameter context.

\subsection{Training Objective}
To train \proj, we minimize the mean absolute percentage error (MAPE), a scale-invariant metric suited for the wide dynamic ranges often found in analog designs. Given ground-truth values $y_k$ and predictions $\hat{y}_k$:
\begin{equation}
\mathcal{L}_{MAPE} = \frac{1}{K}\sum^K_{k=1}\frac{|y_k - \hat{y}_k|}{|y_k|}
\end{equation}
During training, the parameters of \proj are optimized end-to-end by gradient descent. By directly relating the learned representations to performance metrics, the model is guided to capture both circuit structures and parametric dependencies in a manner that enhances predictive accuracy. In practical applications, \proj can be further fine-tuned on specific unseen circuits to  enhancing its adaptability. 

\section{Experiments}

In this section, we detail our experimental setup, dataset preparation, model configurations, and training procedures for evaluating \proj on in-distribution and zero-shot analog circuit performance prediction.

\subsection{Dataset Collection}
Our experiments build on a comprehensive analog amplifier dataset, a domain of critical importance in practical circuit design due to the intricate trade-offs among performance metrics and real-world constraints~\cite{lyu2018batch,settaluri2020autockt}.

\begin{table}[!t]
\centering
\caption{Distribution of Circuits by Number of Devices.}
\label{tab:circuits}
\begin{tabular}{|c|c|c|c|c|c|}
\hline
\# Devices & $<10$ & $10\text{-}14$ & $15\text{-}19$ & $20\text{-}29$ & $>30$ \\
\hline
\# Circuits & $3$ & $18$ & $15$ & $17$ & $7$\\
\hline
\end{tabular}
\end{table}

To comprehensively cover this design complexity, we start with the publicly available AnalogGym~\cite{li2024analoggym} dataset, which includes 16 amplifier topologies. We substantially expand this dataset by collecting and designing an additional set of 44 amplifier topologies, resulting in a total of 60 distinct analog amplifier circuit topologies as shown in Table~\ref{tab:circuits}. For each of these 60 topologies, we perform extensive SPICE-level simulations following the testbench provided by AnalogGym to generate a diverse and large-scale dataset of design instances. Specifically, we systematically vary device parameters, including transistor dimensions (e.g., length \(L\), width \(W\), and multiplicity \(M\)), resistor values, capacitor values and bias currents. We use the Sky130 PDK, with the specific range for these parameters as follows:
\begin{equation}
\begin{split}
W\in[0.2,10]\mu m,\ L\in[0.13,1]\mu m,\ M\in[1, 100], \\ 
C\in[1, 100]pF,\ R\in[0.1, 1000]K\Omega,\ I_{bias}\in[1, 40]\mu A \end{split}
\end{equation} 

Each simulated design instance is characterized by a complete set of critical amplifier performance metrics, summarized in Table~\ref{tab:metrics}. This rich set of performance metrics captures both static and dynamic behaviors essential for realistic analog circuit modeling.


\begin{table}[!t]
\centering
\caption{Statistics of Metrics for Op-Amps in the Training Dataset}
\label{tab:metrics}
\begin{tabular}{|c|c|}
\hline
Metrics & Median\\
\hline
Power Consumption (mW) & 0.2105 \\
DC Gain (dB) & 123.0 \\
Gain-Bandwidth Product (MHz) & 1.182 \\
Phase Margin (\textdegree) & 69.03 \\
Slew Rate (V/$\mu$s) & 0.4425 \\
Settling Time ($\mu$s) & 8.732 \\
CMRR DC (dB) & -90.26 \\
Positive PSRR (dB) & -91.12 \\
Negative PSRR (dB) & -92.52 \\
Input Offset Voltage (mV)  & 0.05050 \\
Temperature Coefficient (ppm/\textdegree C) & 0.5791 \\
\hline
\end{tabular}
\end{table}

We simulate 60,000 parameter-performance pairs for each topology, culminating in a substantial dataset of 3.6 million design instances. To evaluate \proj’s ability to generalize both within familiar topologies and across entirely new circuit structures, we partition the dataset strategically. We divide the topologies into three subsets: training (80\%), validation (10\%), and testing (10\%). Besides, we reserve 10 unseen amplifier topologies exclusively for the testing set, providing a realistic and challenging zero-shot evaluation scenario. In this setting, the model must predict performance metrics for circuits structurally distinct from any it has previously encountered, thus rigorously testing its generalization capabilities.

\subsection{Experimental Setup}
\subsubsection{Model configuration}

\proj employs a transformer-based architecture tailored for hierarchical information processing in analog circuit graphs. Our final model configuration consists of 6 encoder layers and 2 decoder layers, with a hidden embedding dimension of 512, 8 attention heads, and a feed-forward dimension of 2048. To effectively capture global circuit information, we introduce a dedicated graph-level token alongside the detailed pin-level representations. We start the adaptive parameter injection from the fourth layer of the encoder. Dropout of 0.1 is applied to all linear and attention projections for robust generalization.

\subsubsection{Training and evaluation}

We train \proj using the Adam optimizer with an initial learning rate of \(5\times 10^{-4}\), applying a cosine decay schedule and gradient clipping. During training, we normalize the dataset using the mean and standard deviation computed from the training set for each performance metric. The model is trained for 200 epochs with 256 batch size on 2$\times$ A100 GPUs. We use MAPE as our loss function due to its suitability for capturing relative prediction errors across performance metrics spanning diverse scales.

For evaluation, we also report the Acc@K metric~\cite{hakhamaneshi2022pretraining}, which measures the ratio of predictions within a certain precision w.r.t to the ground truth values:
\begin{equation}
Acc@K = \frac{1}{N}\sum^N_{j=1}\mathbbm{1}(|y_j - \hat{y}_j| \le \frac{max_iy_i - min_iy_i}{K})
\end{equation}

\subsection{Comparisons with Baselines}
We compare \proj against several widely adopted baselines for analog circuit performance prediction. These baselines span non-structural MLP, GNNs, and transformer-based architecture, offering a comprehensive comparison of different modeling capabilities.

We first implement a simple MLP baseline, which processes serialized representations of circuit topologies. Device parameters with multiple dimensions are encoded into a unified embedding vectors and concatenated with the corresponding device and pin embeddings. However, the MLP inherently lacks the ability to explicitly represent circuit connectivity, limiting its effectiveness in capturing topological interactions. The  GCN~\cite{kipf2016semi} baseline is a vanilla GNN architecture, proven effective for sizing tasks~\cite{wang2020gcn, li2023design}. We implement the GCN backbone using the same learnable encoding strategy for pin-level nodes and directly inject the parameter embeddings into node features. To explore more expressive GNN variants, We adapt DeepGEN~\cite{li2020deepergcn}, an advanced architecture that incorporates residual connections and deeper aggregation layers. DeepGEN has demonstrated superior performance over conventional GNNs in circuit-related applications~\cite{hakhamaneshi2022pretraining}, and we modify it to align with our training pipeline for a fair comparison.
In contrast to GNN-based models, we also include the GTN~\cite{fan2024graph}, which leverages attention mechanisms to model dependencies across circuit graphs. For fair comparison, all baseline models and \proj share identical transformer decoder structures and identical depth configurations, ensuring consistency in model complexity and representational capacity.

The models are evaluated under two distinct scenarios: (1) In-distribution, where circuit topologies are involved during the training phase but tested under novel parameter configurations; and (2) Zero-shot, where the models are tested on entirely new circuit topologies. Each test circuit structure includes 6,000 parameter-performance pairs, enabling balanced and robust evaluations. 

\subsubsection{Zero-shot performance}
As summarized in Table~\ref{tab:performance_comparison}, the simple MLP baseline exhibits significantly poorer zero-shot performance due to its inability to capture circuit connectivity explicitly. In contrast, both GCN and DeepGEN demonstrate improved generalization to unseen topologies, benefiting from the structural representation capabilities inherent to GNN architectures. GTN achieves slightly better performance than DeepGEN, likely due to its attention-based architecture, which effectively models long-range dependencies across the circuit graph. Among all evaluated models, \proj consistently achieves superior performance across both MAPE and Acc@K metrics. Notably, in the zero-shot setting, \proj significantly reduces prediction errors and improves accuracy, highlighting its strong generalization capability to novel circuit structures.

\begin{table}[!t]
    \setlength{\tabcolsep}{4pt}
    \centering
    \caption{Performance Comparison of Different Models for Trained and Zero-shot Scenarios}
    \begin{tabular}{lcccc}
        \toprule
        \multirow{2}{*}{Model} & \multicolumn{2}{c}{MAPE($\downarrow$)} & \multicolumn{2}{c}{Acc@K($\uparrow$)} \\
        & In-distribution & Zero-shot & In-distribution & Zero-shot \\
        \midrule
        MLP & 0.153 & 0.451 & 0.301 & 0.015\\
        GCN~\cite{wang2020gcn} & 0.103 & 0.256 & 0.685 & 0.367\\
        DeepGEN~\cite{hakhamaneshi2022pretraining} & 0.092 & 0.214 & 0.732 & 0.493\\
        GTN~\cite{fan2024graph} & 0.095 & 0.192 & 0.758 & 0.542   \\
        Ours & \textbf{0.090} & \textbf{0.143} & \textbf{0.865} & \textbf{0.645}\\
        \bottomrule
    \end{tabular}
    \label{tab:performance_comparison}
\end{table}

The results validate the effectiveness of our hierarchical information processing framework and the progressive knowledge fusion mechanisms integrated into \proj. Overall, \proj emerges as a scalable and reliable solution for zero-shot performance modeling, effectively reducing dependence on computationally expensive circuit simulations.

\subsubsection{Few-shot performance}
\begin{figure}[!t]
\centering
\includegraphics[width=0.8\columnwidth]{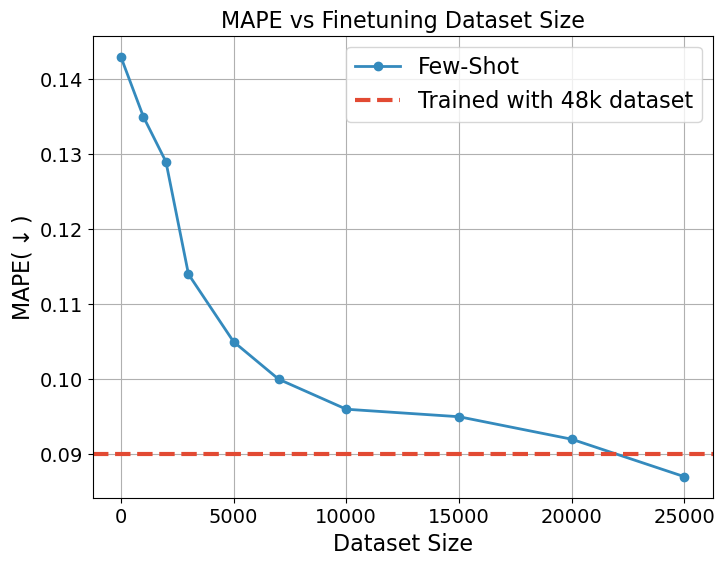}
\caption{Comparison of few-shot learning accuracy across different fintuning dataset sizes against fully trained circuit topology baseline.}
\label{fig:few_shot}
\end{figure}

We further evaluate the adaptability of \proj under few-shot learning settings, evaluating its ability to generalize to novel circuit topologies with limited fine-tuning data. we progressively scale the fine-tuning dataset size from 1,000 to 25,000 parameter-performance pairs using the same unseen topologies as in the zero-shot experiments. 
As illustrated in Fig.~\ref{fig:few_shot}, \proj exhibits rapid performance improvements with increasing fine-tuning data size. Remarkably, with only 20,000 fine-tuning instances, \proj achieves predictive performance comparable to that of fully trained models on in-distribution topologies. This demonstrates its ability to quickly adapt to new designs with limited additional training data, further highlighting its utility for practical analog circuit design applications.

\subsection{Ablation Study}
To thoroughly evaluate \proj’s design choices, we perform ablation studies spanning the encoder backbone and decoder design, identifying key architectural contributions and their impact on accuracy and generalization.

\subsubsection{Analysis of different components}

\begin{table*}[!t]
\caption{Ablation study of key architectural components in \proj. We evaluate the impact of the graph token, context-level processing, and adaptive parameter injection. Results are reported for both trained and zero-shot settings using MAPE($\downarrow$) and Acc@K($\uparrow$).}
\centering
\begin{tabular}{l|ccc|cc|cc}
\toprule
\multirow{2}{*}{Model} & \multirow{2}{*}{Graph Token} & Context & Adaptive Parameter & \multicolumn{2}{c|}{MAPE($\downarrow$)} & \multicolumn{2}{c}{Acc@K($\uparrow$)} \\
 & & Knowledge &Injection &In-distribution & Zero-shot & In-distribution & Zero-shot\\
\midrule
Baseline & \ding{55} & \ding{55} & \ding{55} & 0.107 & 0.312 & 0.698 & 0.406 \\
Graph Token & \ding{51} & \ding{55} & \ding{55} & 0.099 & 0.201 & 0.754 & 0.487 \\
Context-Aware & \ding{51} & \ding{51} & \ding{55} & 0.093 & 0.178 & 0.842 & 0.546 \\
\proj & \ding{51} & \ding{51} & \ding{51} & \textbf{0.090} & \textbf{0.143} & \textbf{0.865} & \textbf{0.645} \\
\bottomrule
\end{tabular}
\label{tab:components}
\end{table*}

We first analyze key components in the encoder backbone: the graph token, context-aware hierarchical information processing, and adaptive parameter injection. Each ablated variant removes or disables one component while keeping the rest of the architecture unchanged. The baseline represents a standard transformer encoder that operates on adjacency-based graph representations.

As shown in Table~\ref{tab:components}, introducing a graph token improves generalization. By enabling global aggregation across the entire circuit, it leads to an improvement in zero-shot accuracy (with Acc@K increases from 0.406 to 0.487), demonstrating its effectiveness in capturing high-level circuit semantics. Incorporating hierarchical structure via alternating structure-refining and context-enhancing transformer blocks further boosts performance. The model becomes more aware of both local and global circuit interactions. Zero-shot MAPE improves from 0.201 to 0.178, while Acc@K rises from 0.487 to 0.546, validating the effectiveness of multi-scale representation learning. Finally, enabling adaptive parameter injection via cross-attention with device-specific masks preserves the consistency of the initial graph representation while injecting the model with parametric information. This addition yields the largest performance gain in the zero-shot scenario, improving MAPE from 0.178 to 0.143 and Acc@K from 0.546 to 0.645. These results demonstrate the critical role of this mechanism in enabling generalization to new topologies with various parameter distributions.

In summary, each component plays a complementary role in enhancing performance. The full \proj architecture integrating all three components, achieves the best results across all metrics. 
This highlights the importance of jointly modeling structural, contextual, and appropriate parametric information within a unified transformer framework to achieve robust and scalable zero-shot circuit performance prediction.

\subsubsection{Analysis of decoder}

\begin{table}[h]
\centering
\caption{Ablation Study of MAPE for different decoder structures}
\begin{tabular}{lcc}
\toprule
Model Variant & In-distribution & Zero-shot \\
\midrule
MLP (Graph Token) & 0.104 & 0.223 \\
MLP (Pooling Embedding) & 0.107 & 0.276\\
Transformer Decoder & \textbf{0.090} & \textbf{0.143}  \\
\bottomrule
\end{tabular}
\label{tab:decoder}
\end{table}

We further explore the impact of different decoder designs on model performance. Specifically, we compare our query-based transformer decoder with two alternatives: MLP (Graph Token), which uses only the learned graph token embedding as input to a multilayer perceptron for final performance prediction; MLP (Pooling Embedding), where we compute a global circuit embedding via average pooling over all node representations and pass the resulting vector through an MLP for prediction.

As shown in Table~\ref{tab:decoder}, both MLP-based approaches underperform relative to the transformer decoder. The performance gap is particularly evident in the zero-shot setting, where modeling complex cross-node and parameter interactions becomes critical. The transformer decoder, with its attention-based query mechanism, effectively integrates structural and parametric information, enabling superior generalization to unseen topologies. This highlights the necessity of flexible and expressive decoding architectures for accurate analog performance modeling.


\subsubsection{Embedding visualization}
\begin{figure}[!t]
\centering
\includegraphics[width=0.7\columnwidth]{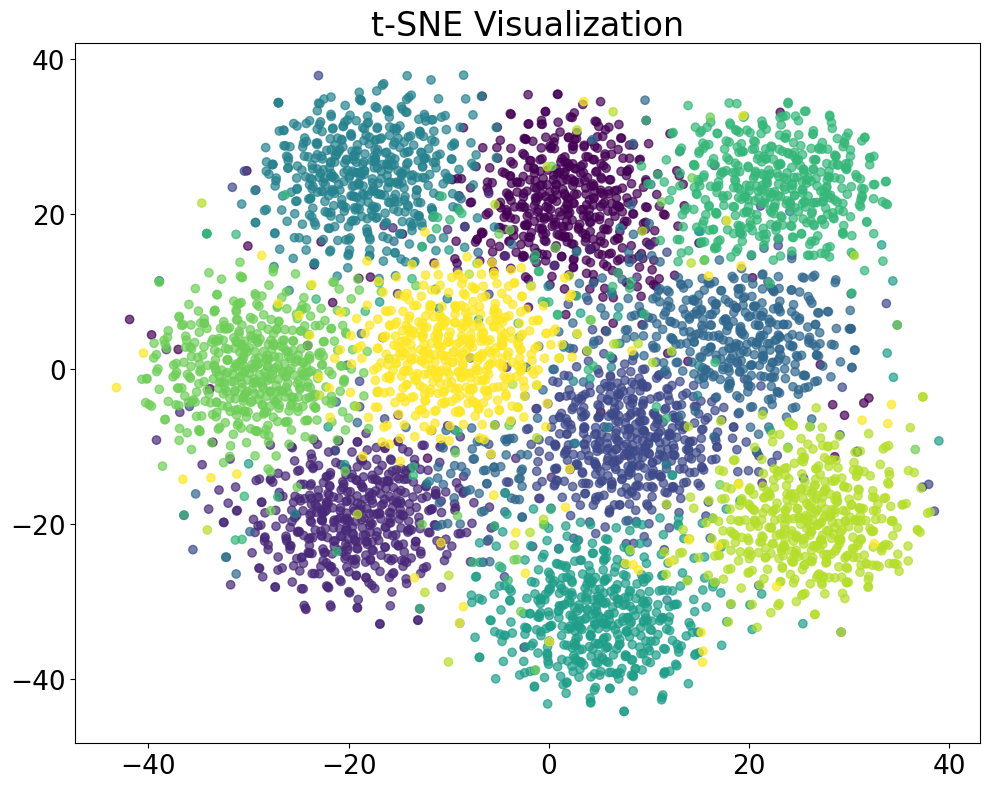}
\caption{t-SNE visualization of graph embeddings from the \proj encoder for 500 randomly selected samples across 10 circuit topologies. Each color represents a different topology.}
\label{fig:tsne}
\end{figure}

To better understand how our model encodes circuit topology and parameter information, we visualize the learned embeddings using t-SNE. In this analysis, we randomly sample 500 data points from 10 distinct circuit topologies. Each embedding is extracted from the encoder's output representation. As illustrated in Fig.~\ref{fig:tsne}, the embeddings form well-separated clusters, with each cluster corresponding to a unique circuit topology. This clustering behavior indicates that \proj effectively captures structural consistency across samples from the same topology, despite differences in device parameters. The results demonstrate the model’s capability to embed circuits into a unified latent space, where structurally similar designs remain close even when parameters vary, highlighting the strength of our encoder in supporting both generalization and discriminative representation learning.

\subsection{RL-based Optimization}
\begin{figure}[!t]
\centering
\includegraphics[width=1\columnwidth]{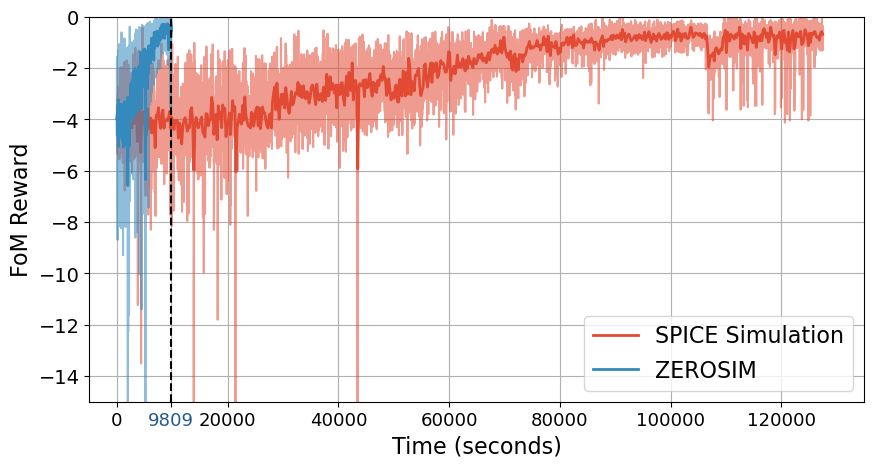}
\caption{Performance comparison of \proj and SPICE simulation when adopted in the RL-based circuit sizing pipeline. The lines are averaged over 200-second window. \proj's circuit parameters are re-evaluated by SPICE simulation for a fair FoM reward comparison. Reward is defined in AnalogGym~\cite{li2024analoggym}.}
\label{fig:rl}
\end{figure}

We evaluate the effectiveness of RL-based circuit sizing pipelines across different circuit evaluation methods. Specifically, we adopt the RL pipeline from AnalogGym~\cite{li2024analoggym, li2023design} with both traditional SPICE simulation and our proposed \proj on a zero-shot circuit (NMCF), which is excluded from the training set. Following AnalogGym, a figure-of-merit (FoM) reward of 0 indicates that the current parameters meet the design goals. As shown in Fig.~\ref{fig:rl}, the RL sizing process is accelerated by nearly 13$\times$ using \proj instead of traditional SPICE simulation to reach the 0 FoM. The final parameter configuration obtained via \proj is further validated through SPICE simulation, confirming that the design requirements are indeed satisfied. This result demonstrates the practical use of our method.

\section{Conclusion}
In this paper, we introduced a unified transformer-based framework, \proj, for zero-shot analog amplifier circuit performance evaluation. By leveraging extensive training data and a unified embedding strategy incorporating global circuit characteristics and adaptive parameter injection, \proj generalizes across unseen amplifier topologies without additional retraining. Experimental evaluations demonstrate superior accuracy compared to existing methods and achieve significant speedup when integrated into reinforcement learning-based sizing optimization workflows. However, our method requires more computational resources than lightweight proxy models designed for simple single-circuit prediction tasks. Moreover, its current validation is limited to amplifier circuits. In future work, we plan to extend the model’s applicability to a broader range of circuit types and explore its integration into fully automated circuit synthesis pipelines.

\section*{Acknowledgment}
This work was partially supported by NSF Award \#2526432.

\bibliography{reference}
\bibliographystyle{IEEEtran}

\end{document}